\documentclass[sigconf, 9pt]{acmart}

\usepackage{times}  
\usepackage{hyperref}
\usepackage{color}
\usepackage{xspace}
\usepackage{amsmath}
\usepackage{todonotes}
\usepackage{subfigure}
\usepackage{url}
\usepackage{booktabs}     

\newcommand{\eg}{e.g., }
\newcommand{\ie}{i.e., }

\newcommand\myeq{\mathrel{\overset{\makebox[0pt]{\mbox{\normalfont\tiny\sffamily def}}}{=}}}

\hypersetup{pdfstartview=FitH,pdfpagelayout=SinglePage}

\newcommand{\flaas}{{\sf{FLaaS}}\xspace}

\newif\ifcomment
\commenttrue

\ifcomment
    \newcounter{DPNOTENumberOfComments}
    \stepcounter{DPNOTENumberOfComments}
     \newcommand{\dpnote}[1]{\textcolor{green}{\small \bf [dp\#\arabic{DPNOTENumberOfComments}\stepcounter{DPNOTENumberOfComments}: #1]}}
    
\else
    \newcommand\dpnote[1]{}
\fi

\begin{document}

\acmConference[DistributedML'20]{1st Workshop on Distributed Machine Learning}{Dec. 1, 2020}{Barcelona, Spain}
\acmBooktitle{1st Workshop on Distributed Machine Learning (DistributedML'20), Dec. 1, 2020, Barcelona, Spain}
\acmPrice{15.00}
\acmDOI{10.1145/3426745.3431337}
\acmISBN{978-1-4503-8182-6/20/12}

\copyrightyear{2020}
\setcopyright{acmlicensed}
\acmYear{2020}


\title[FLaaS: Federated Learning as a Service]{FLaaS: Federated Learning as a Service}

\author{Nicolas Kourtellis}
\affiliation{
    \institution{Telefonica Research}
    \city{Barcelona}
    \country{Spain}
}
\email{nicolas.kourtellis@telefonica.com}

\author{Kleomenis Katevas}
\affiliation{
    \institution{Telefonica Research}
    \city{Barcelona}
    \country{Spain}
}
\email{kleomenis.katevas@telefonica.com}

\author{Diego Perino}
\affiliation{
    \institution{Telefonica Research}
    \city{Barcelona}
    \country{Spain}
}
\email{diego.perino@telefonica.com}

\begin{abstract}
Federated Learning ($FL$) is emerging as a promising technology to build machine learning models in a decentralized, privacy-preserving fashion.
Indeed, $FL$ enables local training on user devices, avoiding user data to be transferred to centralized servers, and can be enhanced with differential privacy mechanisms.
Although $FL$ has been recently deployed in real systems, the possibility of collaborative modeling across different 3rd-party applications has not yet been explored.
In this paper, we tackle this problem and present Federated Learning as a Service (\flaas), a system enabling different scenarios of 3rd-party application collaborative model building and addressing the consequent challenges of permission and privacy management, usability, and hierarchical model training.
\flaas can be deployed in different operational environments.
As a proof of concept, we implement it on a mobile phone setting and discuss practical implications of results on simulated and real devices with respect to on-device training CPU cost, memory footprint and power consumed per $FL$ model round.
Therefore, we demonstrate \flaas's feasibility in building unique or joint $FL$ models across applications for image object detection in a few hours, across 100 devices.
\end{abstract}

\maketitle

\section{Introduction}

Machine Learning as a Service (MLaaS) has been on the rise in the last years due to increased collection and processing of big data from different companies, availability of public APIs, novel advanced machine learning (ML) methods, open-sourced libraries, tools for large-scale ML analytics and cloud-based computation.
Such MLaaS systems have been predominantly centralized: all data of users/clients/devices need to be uploaded to the cloud service provider (e.g., Amazon Web Services~\cite{aws-mlaas2020}, Google Cloud~\cite{googlecloud-mlaas2020} or Microsoft Azure~\cite{azure-mlaas2020}), before ML data modeling can be done.

Federated Learning ($FL$)~\cite{konecny2016federated-learning} is a natural evolution of centralized ML methods, as it allows companies employing $FL$ to build ML models in a decentralized fashion close to users' data, without the need to collect and process them centrally.
In fact, $FL$ has been extended and applied in different settings, and recently deployed in real systems (e.g., Google Keyboard~\cite{bonawitz2019FL-sysml}).
Also, privacy guarantees can be provided by applying methods such as Differential Privacy (DP)~\cite{geyer2017FL-DP-central}, or by computing models within a P2P network~\cite{bellet2018p2p-ml,roy2019brain-torrent}.
In addition, recent start-up efforts in $FL$ space (e.g.,~\cite{openmined2020, datafleets2020, DML2018}) aim to provide $FL$ support and tools to 3rd-party customers or end-users.
However, these solutions have two shortcomings.
First, they do not allow independent 3rd-parties to collaborate and build common ML models while protecting users' privacy.
This feature would enable different applications to collaborate to build better models thanks to larger and richer datasets, while preserving users' privacy.
Second, they do not provide an ``as a service model'' like existing MLaaS platforms.
This feature is critical to enable developers and data scientists to deploy quickly and easily $FL$ solutions, also fostering large adoption of the $FL$ paradigm.

To enable these two features, there are a few fundamental challenges in $FL$ space to be addressed first:
\begin{itemize}
    \item \textit{How do we enable collaborative modeling across different 3rd-party applications, to solve existing or new ML problems?}
    \item \textit{How do we perform effective permission and privacy management of data and models shared across collaborating parties?}
    \item \textit{How do we take advantage of topological properties of communication networks, for better $FL$ modeling convergence, without compromising user and data privacy?}
    \item \textit{How do we provide collaborative $FL$ models in an ``as a Service'' fashion?}
\end{itemize}

\begin{figure*}[t]
\centering
    \includegraphics[scale=0.6]{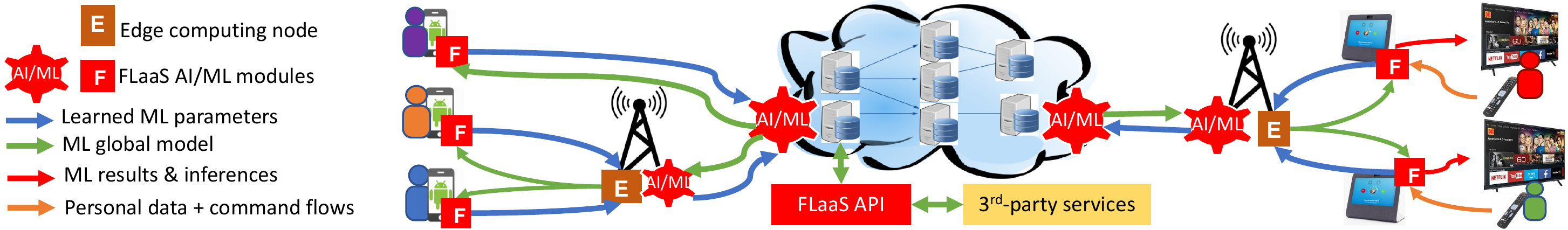}
    \vspace{-3mm}
    \caption{Examples of use cases to be supported by FLaaS. Applications can request FL-modeling support while executed on mobile devices (left-side), or IoT and home devices like personal assistants and smart TVs (right-side).
    Also, FLaaS can employ edge computing nodes to improve ML convergence, but not compromising users' privacy.}
\label{fig:use-cases}
\vspace{-4mm}
\end{figure*}

In this paper, we propose the \textit{Federated Learning as a Service} (\flaas), to address such challenges and facilitate a wave of new applications and services based on $FL$.
In particular, \flaas makes the following contributions in $FL$ space:
\begin{enumerate}
    \item provides high-level and extensible APIs, and an SDK for service usage and privacy/permissions management;
    \item enables the collaborative training of ML models across its customers on the same device using said APIs, in a federated, secured, and privacy-preserving fashion;
    \item enables the hierarchical construction and exchange of ML models across the network;
    \item can be instantiated in different types of devices and operational environments: mobile phones, home devices, edge nodes, etc. (cf.  Figure~\ref{fig:use-cases});
    \item provides the first, to our knowledge, experimental investigation of on-device training costs of FL modeling on actual mobile devices.
\end{enumerate}

Independently from the operational environment, in Section~\ref{sec:motivation}, we present different ways that \flaas supports the building of collaborative models across 3rd-party applications in its $FL$ environment, along with the challenges \flaas must address.
Then, we detail the \flaas system design, APIs and software libraries and how \flaas supports collaborative modeling, in Section~\ref{sec:design}.
As a proof of concept, in Section~\ref{sec:experiments}, we deploy \flaas on a mobile phone setting to assess the practical overheads of running such a service on mobiles.
We demonstrate \flaas capabilities in building unique or joint $FL$ models for image object detection, for independent or collaborative mobile apps using shared data, respectively.
We measure data sharing, ML training and evaluation costs of \flaas with respect to CPU utilization, memory, execution time and power consumption, for on-device $FL$ modeling in the above scenarios.
We show that \flaas can build $FL$ models on mobile phones over 10s of rounds in a few hours, across 100 devices, with 10-20\% CPU utilization, 10s of MBs average memory footprint and 3-5\% battery consumption, per $FL$ round and ML model trained.

\section{Background and Related Work}
\label{sec:background}

Here, we first cover fundamental assumptions on $FL$ and the federated optimization problem, and then academic or industrial efforts on the topic of distributed $ML$ and $FL$.

\subsection{Preliminaries on Federated Learning}
\label{sec:fl-preliminaries}

\noindent
\textbf{Assumptions.} The $FL$ optimization problem has the following typical assumptions~\cite{konecny2016federated-learning,mcmahan2017-FL}:
\begin{itemize}
\setlength\itemsep{-0pt}
    \item \textit{Massively distributed:} Number of examples (data) available per client (device) is expected to be much smaller than the number of clients participating in an optimization.
    \item \textit{Limited Communication:} Devices can be typically offline or with intermittent or slow connectivity; their bandwidth can be considered expensive commodity (especially if no WiFi connection is available).
    \item \textit{Unbalanced:} Data available for training on each device is of different sizes, since users can have different usage profiles (some can be heavy hitters, others on the tail~\cite{zhu2020heavyhitters}).
    \item \textit{Highly Non-IID:} Data available on each client is typically non representative of the population distribution, as they only reflect the given client's usage of the device.
    \item \textit{Unreliable Compute Nodes:} Devices can go offline unexpectedly; there is expectation of faults and adversaries.
    \item \textit{Dynamic Data Availability:} The subset of data available is non-static, e.g., due to differences in hour, day, country.
\end{itemize}

\noindent
\textbf{Notations.}
The next analytical formulations use these notations:
\vspace{-0pt}
\begin{itemize}
\setlength\itemsep{-0pt}
    \item Set of devices $\mathbb{K}$; total number of devices $K=|\mathbb{K}|$;
    \item A given client is identified as $k \in \mathbb{K}$;
    \item Total number of rounds $R$;
    \item A given round is identified as $r \in [0,R]$;
    \item Fraction of devices used per round: $C \in [0,1]$;
    \item Number of data samples across all clients: $n$;
    \item Set of indices of data samples on $k$: $P_k$;
    \item Number of data samples available at $k$: $n_k = |P_k|$;
    \item Batch size of data samples used per client: $B$;
    \item Number of iterations of device $k$ on local data: $t \in E$;
\end{itemize}

\noindent
\textbf{Federated Aggregation:}
We consider $FL$-based algorithms with finite-sum objectives of the form:
\vspace{-12pt}
\[
\min\limits_{w \in \mathbb{R}^d} f(w), \quad \text{where} \quad 
f(w)\myeq\frac{1}{n}\displaystyle\sum_{i=1}^{n}f_i(w)
\]
\vspace{-10pt}

\noindent $f_i(w)$ is defined as loss of prediction on seen examples, i.e., $l(x_i;y_i;w)$, using the trained model with parameters $w$.
All data available in the system ($n$) are partitioned over $K$ clients, each with a subset of indices $P_k$.
Then, the problem objective can be rewritten as:
\vspace{-6pt}
\[
f(w)=\displaystyle\sum_{k=1}^{K}\frac{n_k}{n}F_k(w), \quad \text{where} \quad
F_k(w) = \frac{1}{n_k}\displaystyle\sum_{i\in P_k}f_i(w)
\]
Given the assumption of non-IID data, $F_k$ could be an arbitrarily good or bad approximation to $f$.

The model optimization under $FL$ assumes a decentralized architecture, which performs a \textit{Federated Aggregation} algorithm with the clients and a central server.
Each client $k$ executes a stochastic gradient decent ($SGD$) step (iteration $t$) on the local data available, $g_k$$=$$\nabla$$F_k(w_t)$.
Assuming C=1, at iteration $t$, the server aggregates all gradients and applies the update on the global model: $w_{t+1} \leftarrow w_t - \eta \sum^{K}_{k=1}\frac{n_k}{n}g_k$, since $\sum^{K}_{k=1}\frac{n_k}{n}g_k = \nabla f(w_t)$, where $\eta$ is a fixed learning rate.
Equivalently, every client can perform the update as: $w^k_{t+1}$$\leftarrow$$w_t$$-$$\eta$$g_k$, and the global model is $w_{t+1} \leftarrow \sum^{K}_{k=1}\frac{n_k}{n}w^k_{t+1}$.

In fact, this process can be repeated for $t\in E$ iterations locally per client, before sharing models with the server in $R$ rounds.
Therefore, the client can iterate the local update $w^{k} \leftarrow w^{k} - \eta \nabla F_k(w^{k})$ for $E$ times, before the aggregation and averaging at the central server, per round $r$:
$w_{r+1} \leftarrow w_r - \eta \sum^{K}_{k=1}\frac{n_k}{n}g_k$.

It becomes apparent that factors such as $E$ iterations per client, $C$ clients participating in each round, and $R$ rounds executed can have high impact on model performance, and communication cost incurred in the infrastructure to reach it.
We note that $C$ is usually selected in such a way~\cite{bonawitz2019FL-sysml} to account for device unreliability (intermittent connectivity, failed computation, etc.).

\subsection{Related Work}

Several initiatives from startups and online communities have been recently proposed in the space of decentralized ML.
For example, Decentralized Machine Learning (\textit{DML})~\cite{DML2018} is a (now abandoned) blockchain (BC)-based project, enabling its participants to build models in a distributed fashion, while growing its BC network.

Open-source community efforts propose libraries and platforms that will allow users to train ML models in a decentralized, secured and privacy-preserving (PP) fashion.
For example, \textit{OpenMined}~\cite{openmined2020} proposes the libraries \textit{PySyft} and \textit{PyGrid}, by employing multiparty computation (MPC), homomorphic encryption, $DP$ and $FL$ for secured and PP-ML modeling in a decentralized fashion, in both mobile or desktop environments.
In addition, \textit{Datafleets}~\cite{datafleets2020} utilizes $DP$, secure MPC and role-based access control to build models inside or across enterprises, and on edge computing nodes.
With similar technologies, \textit{FATE} (Federated AI Technology Enabler)~\cite{fate2020} focuses on desktop deployments.

Building on the popular \textit{TensorFlow} (TF) framework, \textit{TensorFlow Federated} (TFF) is an open-source framework for ML and other computations on decentralized data~\cite{tensorflow-federated2020}.
\textit{coMind}~\cite{comind2019} proposes a custom optimizer for TF to easily train neural networks via $FL$.
There have also been benchmark frameworks proposed like \textit{LEAF}~\cite{leaf-benchmark2019}, for learning in $FL$ settings, with applications including $FL$, multi-task learning, meta-learning, and on-device learning.

In contrast to all these efforts, \flaas follows a $FL$-as-a-Service model, with high-level APIs, enabling:
a) independent 3rd-party applications (\ie external to \flaas operator) to collaborate and combine their common-type data and models for building joint meta-models for better accuracy;
b) collaborative 3rd-party applications to combine their partial models into meta-models, in order to solve new joint problems, never before possible due to data siloing and applications' isolation;
c) 3rd-party applications to build the aforementioned \flaas models on edge nodes, desktops, mobile phones or other low-resource (IoT) devices (cf. Figure~\ref{fig:use-cases}).

\section{FL\lowercase{aa}S Motivation \& Challenges}
\label{sec:motivation}

\flaas aims at providing to single applications an easy way to use $FL$, without the costly process of developing and tuning the algorithms, as well as to enable multiple applications to collaboratively build models with minimal efforts.
Specifically, \flaas is designed to support the following use cases (some examples in Figure~\ref{fig:use-cases}):

\textbf{1.} Unique $FL$ modeling per individual application for an existing ML problem without the need of developing the algorithms.
Traditionally, $ML$ modeling is requested uniquely per application aiming to solve a specific, existing ML problem: \eg a streaming music application (e.g., Spotify) that wants to model its users' music preferences to provide better recommendations.

\textbf{2.} Unique $FL$ model trained in a joint fashion between two or more collaborative applications for an existing ML problem.
That is, a group $G$ of applications interested in collaborating to build a shared ML model that solves an existing problem, identical and useful for each application, but on more, shared and homogeneous data.
For example, Instagram, Messenger and Facebook (owned by the same company) may want to build a joint ML model for better image recognition, on images of similar quality and scope, but coming from each application's local repository.

\textbf{3.} Unique $FL$ model trained in a joint fashion between two or more collaborative applications, as in case (2), but for a novel, never explored ML problem.
For example, an application for planning your transportation (\eg Uber, GMaps, or Citymapper) may want to model your music preference while on a specific transportation type (\eg bicycle, bus, car, etc.).

\noindent Several challenges arise while supporting these use cases under a Federated Machine Learning setting, which we elaborate next.

\noindent\textbf{Permission management across applications and services:} 
Mobile and IoT systems provide mechanisms to grant application and services access to data such as mobile sensors, location, contacts or calendar. Such access is typically given at a very coarse granularity (\eg~all-or-nothing), and can be unrestricted or, more recently, granted per application. 
On top of these traditional permissions, \flaas has to provide mechanisms to specify permissions across applications and services to share data and models among them.
Further, it has to provide security mechanisms to guarantee these permissions are respected.

\noindent\textbf{Privacy-preserving schemes:} 
In \flaas deployment scenarios and use cases, multiple applications and services can be involved in the $FL$ execution. 
In order to guarantee the privacy of customers' data, it is critical to leverage privacy-preserving mechanisms in the construction of $FL$ models.
In \flaas, we plan to leverage Differential Privacy ($DP$) to provide further privacy guarantees to participating clients.
$DP$ noise can be introduced at different stages of the $FL$ system: in the data source at the client side, also known as local-$DP$~\cite{zhao2020LDP-FL-vehicles,truex2020LDP-FED, seif2020wireless-LDP-FL,wei2020FL-DP-local}, at the central server side~\cite{geyer2017FL-DP-central} while building the global model, or at an intermediate stage such as edge computing nodes~\cite{liu2019cloud-edge-FL} or base stations~\cite{mehdi2019hierarchical-FL-cellular}, or with hybrid methods and hierarchical methods~\cite{zhang2019efficient-FL-DP, truex2019hybrid-FL,briggs2020hierarchical-FL}.
However, introducing $DP$ noise in the ML pipeline reduces model utility~\cite{zhao2020privacy-tradeoff} as it affects convergence rate of the $FL$-trained model.
Note that, while \flaas plans to build on existing DP solutions, finding an optimal way to add $DP$ noise in $FL$ is an open research problem, and beyond the scope of this work. 

\noindent\textbf{Exchange model across a (hierarchical) network with FL:}
As depicted in Figure~\ref{fig:use-cases}, \flaas can build models in a hierarchical fashion across different network layers: end-user device, ISP edge nodes, or the central server.
Recent works considered the hierarchical $FL$ case, where multiple network stages are involved in the training process~\cite{liu2019cloud-edge-FL,mehdi2019hierarchical-FL-cellular,briggs2020hierarchical-FL}.
Such efforts showed convergence and accuracy can be improved with proper design under such settings.
\flaas will build on these works to realize its hierarchical use cases.
However, the design of optimal hierarchical $FL$ methods is an open research problem beyond the scope of this work.

\noindent\textbf{Training convergence and performance:}
As mentioned earlier, the usage of $DP$ in multi-stage training and the hierarchical $FL$ approach impact the convergence and performance of $FL$ models.
However, in \flaas, the possibility of building cross-application models introduces another dimension, potentially impacting model convergence and performance.
This is a relevant research problem that \flaas will need to address in the near future.

\noindent\textbf{Platform usability:}
Every service platform should enable a customer to use its services with limited overhead and knowledge of the underlying technology.
On the one hand, existing commercial MLaaS platforms (\eg AWS~\cite{aws-mlaas2020}, Google Cloud~\cite{googlecloud-mlaas2020} or Azure~\cite{azure-mlaas2020})
provide users with APIs and graphical user interfaces (GUI) to configure and use ML services in cloud environments.
However, these APIs are not designed to deal with cross-application model building, nor tailored for $FL$ services.
On the other hand, existing $FL$ libraries (\eg TFF~\cite{tensorflow-federated2020} or OpenMined~\cite{openmined2020}) are still in prototype phase and cannot support a service model, and do not provide GUIs, or high-level service APIs.
They also do not support cross-application ML modeling as \flaas does.
\flaas builds on these existing works and provides high-level APIs to support model building across applications on the same device and across the network, and software libraries or Software Development Kits (SDKs) for application developers to include the service in their apps and devices (cf. Sec.~\ref{sec:design-api}).

\section{FL\lowercase{aa}S System Design}
\label{sec:design}

\subsection{Service Main Components}
The \flaas design comprises three main system components:

\noindent \textbf{Front-End:} Main interface for customers (e.g., app or service developers), to bootstrap, configure, and terminate the service.
It runs a GUI, processes front-end API calls from customers (or the GUI), and calls functions on the Controller to execute customer requests.

\noindent \textbf{Controller:} Takes as input commands received from Front-End, executes the required steps to configure the service, \eg initialize the model, set appropriate permissions, etc.
Once the service starts, the Controller is in charge of monitoring service health, budget, and terminating execution of ML modeling when requested.

\noindent \textbf{Central Server} and \textbf{Clients} (e.g., mobile or home devices, edge nodes): are the elements actually in charge of executing the $FL$ algorithms and protocol (cf. Sec.~\ref{sec:fl-preliminaries}).
The Central Server, hosting the Controller and Front-End, is under the administrative domain of \flaas, while the Clients are typically in another domain, \eg at user side.
The Server also runs a \flaas Global module responsible for the federated aggregation of received models.
Each Client runs a \flaas Local module directly provided by the \flaas provider.
In addition, every application on the \flaas Clients needs to embed a software library, providing the required functions that can be accessed via client APIs.

\subsection{APIs and Software Libraries}
\label{sec:design-api}

\noindent
\textbf{Front-End APIs:} \flaas can be configured via front-end APIs, or a GUI that uses the front-end APIs under the hood. These APIs can be classified in three types, as follows:

\begin{itemize}
\item \emph{DATA APIs} allow customers to describe data types the model takes as input for training or produces as output after inference.
This is specified via JSON format and includes name and type of each input feature or output data column.
\item \emph{MODEL APIs} enable customers to create an ML model, define model type and parameters, or choose the option of parameter self-tuning.
Also, these APIs allow customers to specify properties in the ML modeling, in case the model is built across (partially) different data from multiple customers, or as an average/ensemble across models.
\item \emph{PERMISSION APIs} enable customers to specify if and which other customers (\eg apps) can access said data, or how other customers can access the model for inference, or to build models to solve new ML problems.
\end{itemize}

\noindent
\textbf{Client APIs:}
A set of functions need to be embedded in the application code to realize \flaas functionality. To this goal, we design a software library (currently implemented as an Android SDK, cf. Sec~\ref{sec:experiments}) providing the implementation of such functions that are then exposed via a set of APIs.
The library includes main functions such as:
(i) Authenticate API to the Central Server.
(ii) Support on-device training for ML applications, including: load a model (either pre-trained or with initial parameters), add training samples, conduct model training, predict from test samples, and save or load model parameters to device memory.
(iii) Exchange/share data between \flaas-enabled apps on-device, before local $FL$ training takes place.

While training on-device a new model is possible, it requires a significant amount of processing power and many communication rounds, making it impractical in user devices with limited resources.
Transfer Learning ($TL$)~\cite{tan2018DL-transfer-learning-survey} is an ML technique that takes a model built for a basic task $T$ (\eg image recognition) and reuses it as the starting point for a new task, but related to $T$, to speed up training and improve model performance.
\flaas employs $TL$ for various scenarios (\eg image object recognition, text classification, item recommendations, etc.) as it is suitable for currently resource-constrained mobile devices and networks.

\subsection{FLaaS Algorithmic Design}
\label{sec:use-cases-algorithms}

\begin{figure}
    \centering
    \includegraphics[scale=0.45]{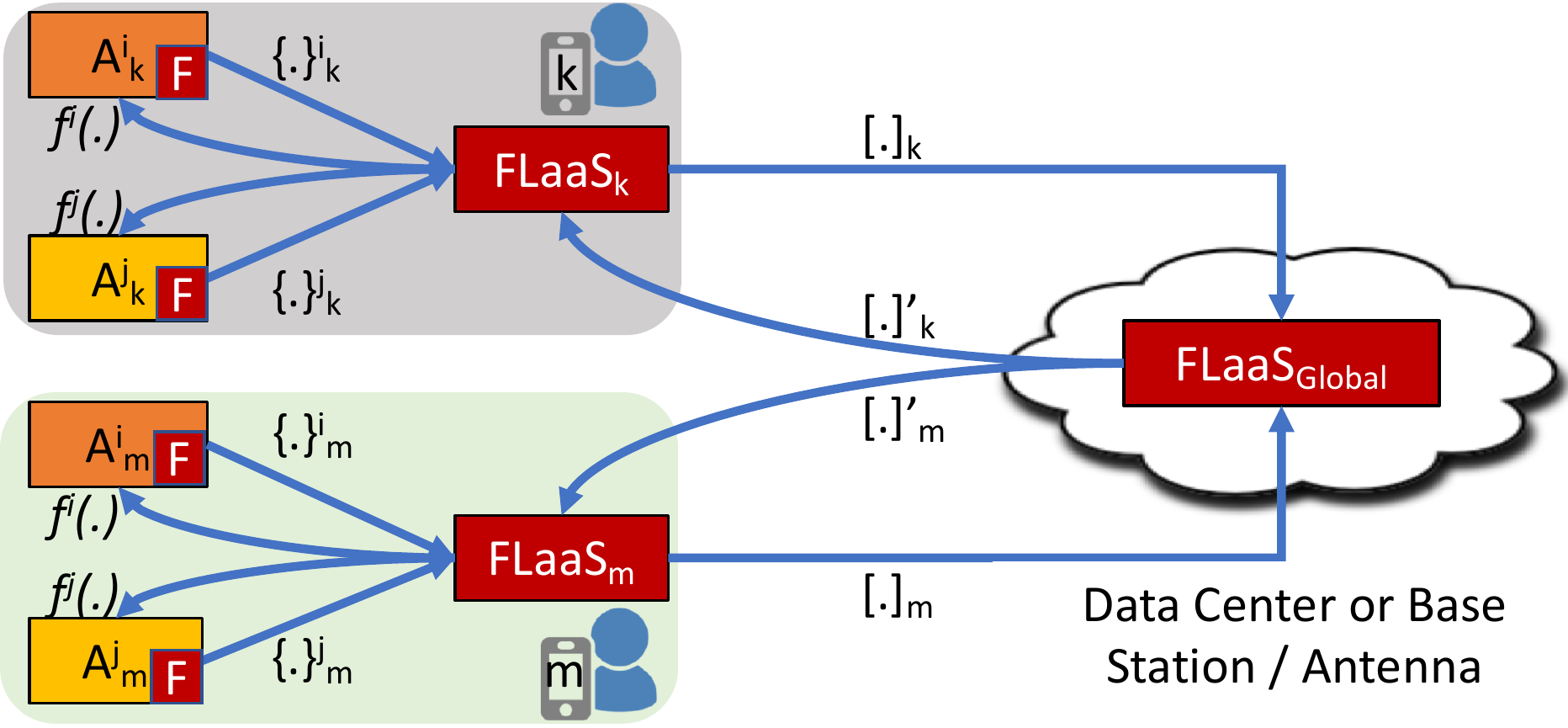}
    \caption{Overview of \flaas ML modeling architecture.}
    \label{fig:architecture}
\vspace{-5mm}
\end{figure}

We now provide algorithmic details of how the main use cases outlined in Sec.~\ref{sec:motivation} are supported by \flaas design.

\noindent
\textbf{FL modeling per application for existing ML problems:} We assume a set of apps, $i \in \mathbb{A}$, installed on device $k \in \mathbb{K}$, are interested in building $FL$ models with \flaas.
In this case, each app wants its own model built on its local data.

Figure~\ref{fig:architecture} outlines the general interactions of two apps $i$ and $j$ with the \flaas Local module running on user devices $k$ and $m$.
The apps communicate to the \flaas Local their $FL$ models built.
Thus, for app $i$ and device $k$, in Fig.~\ref{fig:architecture} we define $\{.\}^i_k = F^i_k(w)$.

\flaas Local collects all such models from individual apps, and transmits them in a compressed format to \flaas Global:
$$[.]_k = [F^i_k(w)], \forall i \in \mathbb{A}, \forall k \in \mathbb{K}$$

Subsequently, \flaas Global performs Federated Aggregation across all reported local models and builds one global weighted average model per app, which it then communicates back to the participating devices per app, \ie in Fig.~\ref{fig:architecture}:
$$[.]'_k = [f^i(w)], \forall i \in \mathbb{A}, \forall k \in \mathbb{K}$$

Finally, the \flaas Local module distributes the global model to each app, \ie in Fig.~\ref{fig:architecture}:
$$f^i(.) = f^i(w), \forall i \in \mathbb{A}$$

\noindent
\textbf{Jointly-trained FL modeling between group of apps for existing ML problem:}
In the following scenario, we assume a group of two or more apps, $i \in \mathbb{A}$, installed on device $k \in \mathbb{K}$, are interested in collaborating and building a common $FL$ model with \flaas.
This model will be shared among all apps but will be built jointly on each application's local data.

Thus, in Figure~\ref{fig:architecture}, we can redefine the general interactions of a group $G$ of apps $i$ and $j$ (\ie $G=\{i,j\}$) with \flaas Local running on devices $k$ and $m$, in order to build such a joint model among them.
In fact, we point out at least three different ways that such joint model can be built, by sharing different elements with \flaas Local (Fig.~\ref{fig:architecture}):

\textit{1. Sharing local data:} $\{.\}^i_k = \{x^i;y^i\}_k$\\
Apps in $G$ share with \flaas Local data they are willing to provide in the collaboration.
\flaas Local collects all shared data, which should have the same format, and performs $SGD$ on them.
For this way to be possible, participating applications must be willing, and permitted to share user data across applications.

\textit{2. Sharing personalized gradients:} $\{.\}^i_k = \{\nabla F^i_k(w_t),\epsilon\}$\\
Apps share with \flaas Local their personalized gradient for iteration $t$ along with the error $\epsilon$, which was acquired after training their local model for the $t^{th}$ iteration.
In this case, \flaas Local uses the received gradients $g^i_k$ per iteration $t$ to incrementally correct the locally built joint model.
Then, it releases back to apps the improved joint model, before receiving new updates in the next iteration.

\textit{3. Sharing personalized model:} $\{.\}^i_k = F^i_k(w)$\\
Apps share with \flaas Local their complete personalized models built on their data, after they perform $E$ iterations.
In this case, \flaas Local performs Federated Aggregation on the received models, thus, building a joint model that covers all apps in $G$, with a generalized, albeit, local model.
In the second and third ways, the apps do not need to worry about permissions for sharing data, as they only share gradients or models.
Note that the user data never leave the device, in any of the aforementioned cases.

Then, for any of these ways, \flaas Local reports to \flaas Global the model jointly built on data or model updates, \ie:
$$[.]_k = [F^G_k(w)], \forall i \in \mathbb{A}, \forall k \in \mathbb{K}$$

Subsequently, \flaas Global performs Federated Aggregation across all collected local models and builds a global weighted averaged model, for each collaborating group $G \in \mathbb{G}$ of applications.
Then, it communicates each such global model back to the participating devices, \ie in Fig.~\ref{fig:architecture}:
$$[.]'_k = [f^G(w)], \forall G \in \mathbb{G}, \forall k \in \mathbb{K}$$

Finally, \flaas Local distributes the global model to each application of the collaborating group $G$, \ie in Fig.~\ref{fig:architecture}:
$$f^i(.) = f^G(w), \forall i \in G$$

\noindent
\textbf{Jointly-trained FL modeling between group of apps for a new ML problem:}
In this scenario, we assume a primary app $i$ is interested in solving a new ML problem but does not have all data required to solve it.
Therefore, it comes to an agreement with other, secondary apps ($j$) to receive such needed data ($x^{j'}$) to build the new model, using \flaas.
Notice that these additional data are a subset of the full data that secondary apps produce, \ie $x^{j'}$$\subseteq$$x^j$.
In a similar fashion as before, the collaborating apps must share data or models, in order to enable joint model building (Figure~\ref{fig:architecture}).
In fact, we point out at least two ways that such joint model can be built, by sharing different elements with \flaas Local:

\textit{1. Sharing local data:}

\indent\indent 1a. Primary application $i$: $\{.\}^i_k = \{x^i;y^i\}_k$

\indent\indent 1b. Secondary applications $j$: $\{.\}^j_k = \{x^{j'}\}_k$

\noindent
Apps share with \flaas Local the data they are willing to provide.
\flaas Local collects all shared data and performs $SGD$ in an iterative fashion, to build the final local model.

\textit{2. Sharing personalized model:}

\indent\indent 2a. Primary application $i$: $\{.\}^i_k = F^i_k(w)$

\indent\indent 2b. Secondary applications $j$: $\{.\}^j_k = F^{j'}_k(w)$

\noindent
Apps provide trained local models that solve portion of the overall new problem, after $E$ iterations.

Then, \flaas Local builds a meta-model (\eg based on hierarchical or ensemble modeling), to solve the new problem at hand.
In either case, again, no data leave the device.
Then, for either of the ways described, \flaas Local reports to \flaas Global the model jointly built, \ie in Fig.~\ref{fig:architecture}:
$$[.]_k = [F^{i'}_k(w)], \forall i' \in \mathbb{A}', \forall k \in \mathbb{K}$$

Note: $\mathbb{A}'$ is the set of primary apps building the novel models and does not include secondary apps helping.
Subsequently, \flaas Global performs Federated Aggregation across collected models and builds a global weighted averaged model for each primary app model requested, and communicates each such global model back to participating devices, \ie in Fig.~\ref{fig:architecture}:
$$[.]'_k = [f^{i'}(w)], \forall i' \in \mathbb{A}', \forall k \in \mathbb{K}$$

Finally, \flaas Local distributes the global model to each primary application, \ie in Fig.~\ref{fig:architecture}:
$$f^{i}(.) = f^{i'}(w), \forall i' \in \mathbb{A}'$$

\section{FL\lowercase{aa}S Proof of Concept}
\label{sec:experiments}

\begin{figure}[t]
    \centering
    \includegraphics[width=1\linewidth]{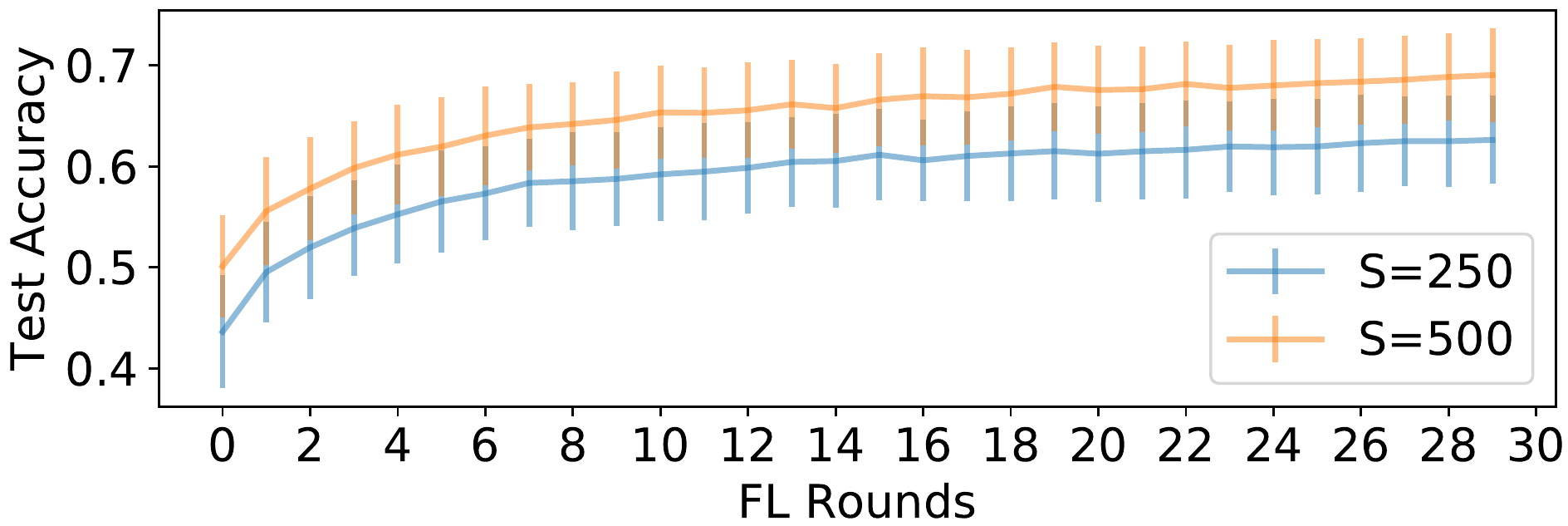}
    \vspace{-7mm}
    \caption{FLaaS test accuracy per FL round, for two sample sizes $S$ available per device. Error bars: accuracy variability for 100 simulated clients.}
    \vspace{-5mm}
    \label{fig:ml-performance}
\end{figure}

\begin{figure*}[t]
\centering
\subfigure[]{
	\includegraphics[width=0.23\linewidth]{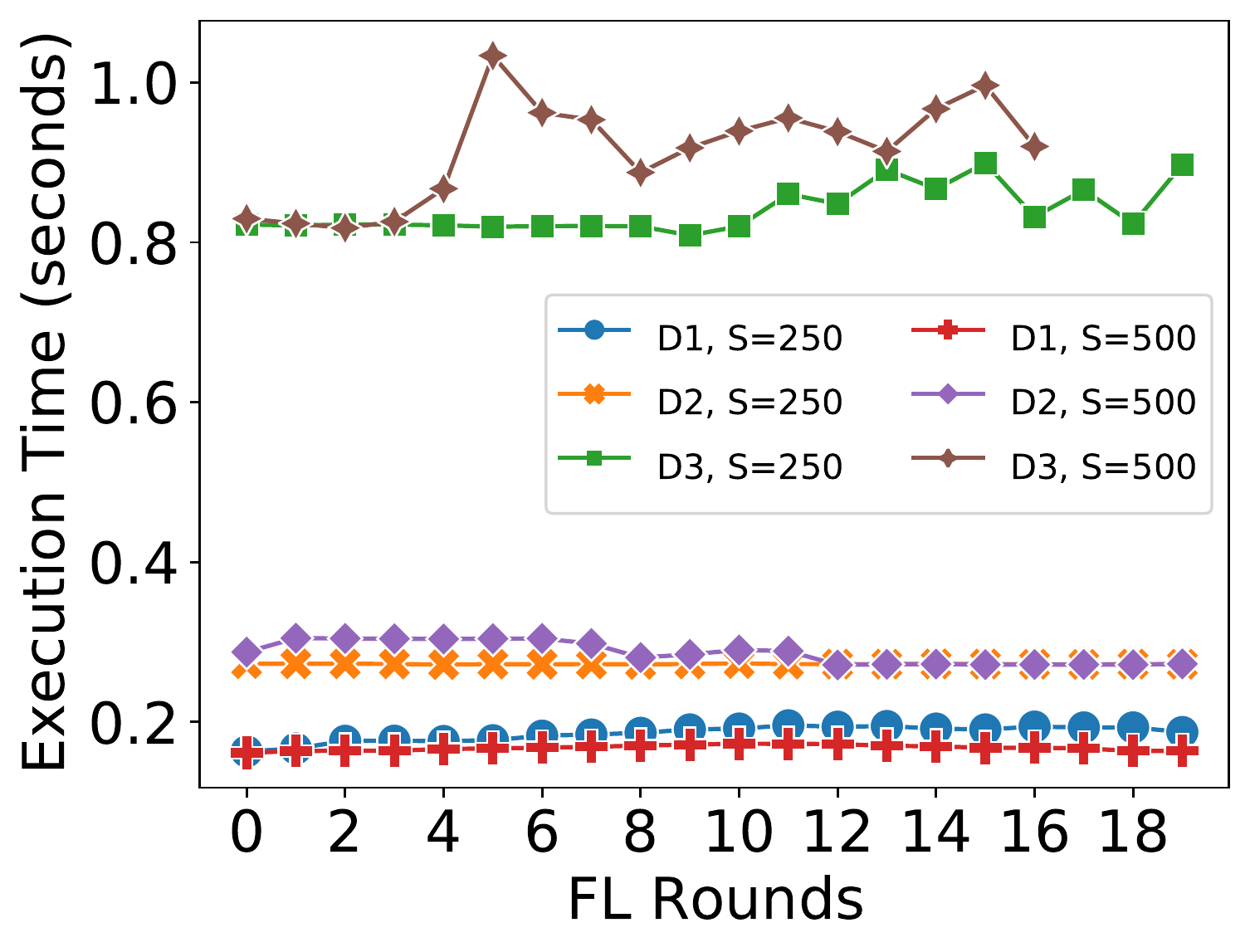}\label{fig:boxplot-time-loadimages}
	}
\subfigure[]{
	\includegraphics[width=0.23\linewidth]{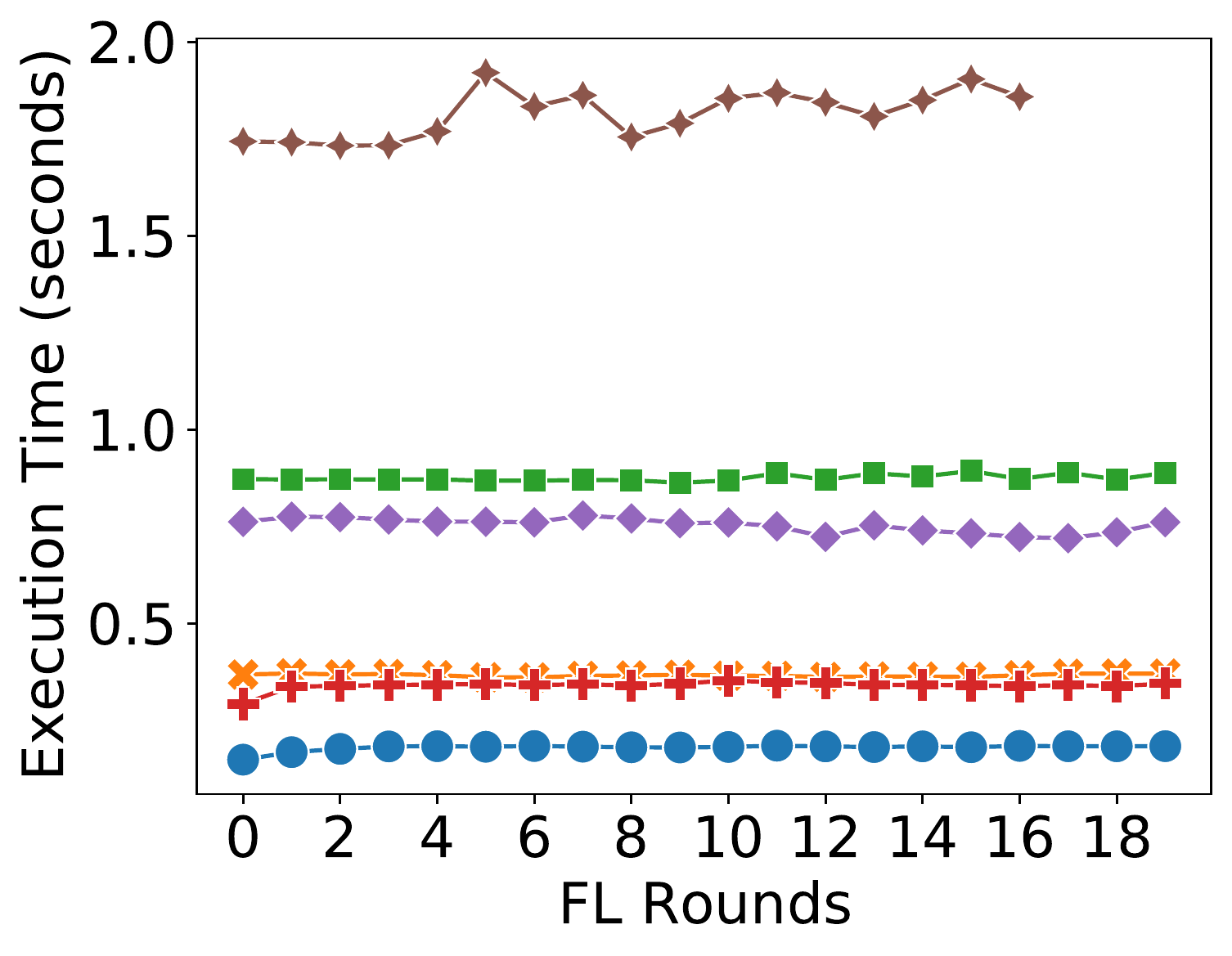}\label{fig:boxplot-time-train}
	}
\subfigure[]{
	\includegraphics[width=0.24\linewidth]{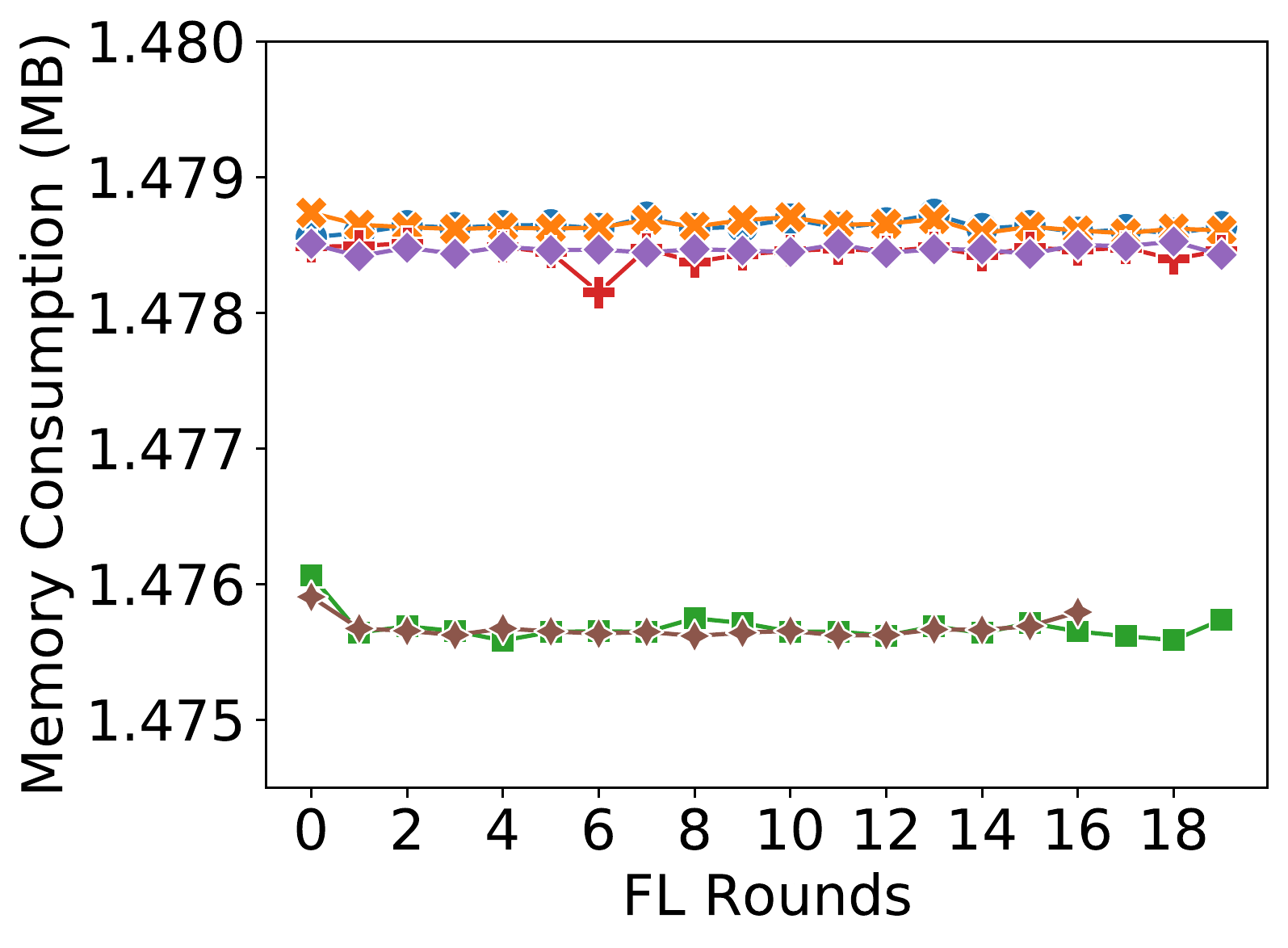}\label{fig:boxplot-memory-loadimages}
	}
\subfigure[]{
	\includegraphics[width=0.24\linewidth]{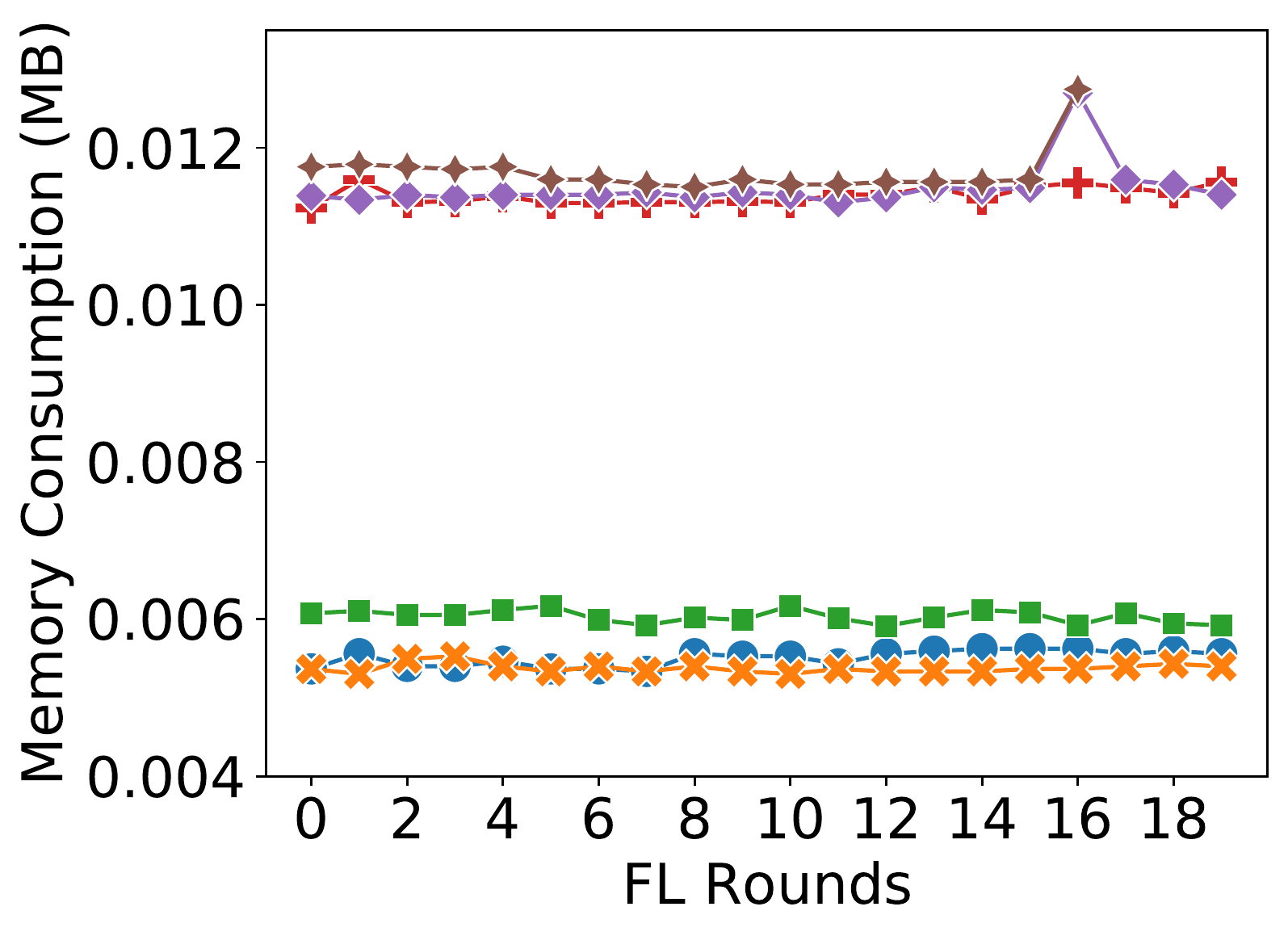}\label{fig:boxplot-memory-train}
	}
	\vspace{-5mm}
	\caption{Average execution time (a) and memory consumed (c) per image loaded for each FL round, and execution time (b) and memory consumed (d) for model training per epoch for each FL round. 3 device types and 2 sample sizes were tested.}
	\vspace{-5mm}
\end{figure*}

We now present a proof-of-concept (PoC) \flaas implementation in the mobile setting.
We discuss early experimental results to showcase viability of two \flaas use cases (Sec.~\ref{sec:use-cases-algorithms}): single app ML modeling and data sharing for ML modeling by two collaborative apps.
 
\noindent
\textbf{PoC FLaaS Implementation.}
On the client side, we focus on Android OS 10 (API 29).
Specifically, we implement the \flaas module as a standalone user level app and the library to be embedded in \flaas-enabled apps as an SDK.
Both \flaas module and SDK leverage TensorFlow Lite 2.2.0 and Transfer API library from Google~\cite{transfer-google}.
Exchange of data between \flaas-enabled apps and the \flaas Local module is performed using the OS's BroadcastReceiver~\cite{broadcastreceiver}.
The Central Server is implemented using Django 3.07 with Python 3.7.
Note that currently, Controller and Front-End are not implemented.
Finally, the \flaas-enabled apps used for the PoC are toy apps only performing the \flaas functionality and storing a set of data.
The current \flaas version is implemented in 4.6k lines of Java code (for the $FL$-related modules), and 2k lines of Python for server modules.

\noindent
\textbf{Experimental Setup.}
We evaluated the performance of our implementation with respect to ML Performance, Execution Time, Memory Consumption, CPU Utilization and Power Consumption using three Android devices.
\emph{D1}: Google Pixel 4 (2.42 GHz octa-core processor with 6GB LPDDR4x RAM). 
\emph{D2}: Google Pixel 3a (2.0 GHz octa-core processor with 4GB LPDDR4 RAM). 
\emph{D3}: Google Nexus 5X (1.8 GHz hexa-core with 2GB LPDDR3 RAM).
We updated all devices to the latest supported OS (D1 \& D2: Android 10; D3: Android 8.1), and disabled automated software updates, battery saver and adaptive brightness features when applicable.
We further set the device under Flight Mode, enabled WiFi access, connected to a stable WiFi 5GHz network, and set the brightness level to minimum.

As a base network, and to initialize the Transfer Learning process, we use MobileNetv2~\cite{s2018mobilenetv2}, pre-trained with ImageNet~\cite{imagenet_cvpr09} dataset with image size 224x224.
As a head network (used for the model personalization), we use a single dense layer, followed by softmax activation (SGD optimizer with 0.003 learning rate). 
As a dataset for model training and testing, we use CIFAR-10~\cite{cifar10}, equally split and distributed across all experimental devices before the experiment.

In our experiments, we applied parameter values used in other $FL$ works with CIFAR-10: 20 samples per batch, 50 epochs, 20 $FL$ rounds, and 250 or 500 samples per user ($S$), corresponding to the two scenarios of individual app or joint $FL$ modeling of two apps via data sharing.
For measuring ML performance, we simulated $FL$ on 100 users (devices) for the two scenarios in 30 $FL$ rounds.
For measuring the on-device cost, we assume 10 users' worth of data and execute $FL$ modeling for the two scenarios on the real devices.

\noindent
\textbf{FLaaS ML Accuracy.}
Figure~\ref{fig:ml-performance} plots the average test accuracy for the two scenarios, per $FL$ round.
These results, acquired using 100 simulated devices, show that when two apps share data ($S$=250+250), 10\% better accuracy can be achieved with the model trained jointly at \flaas Local, than individual models with half data ($S$=250).

\noindent
\textbf{FLaaS On-device Memory \& Execution Costs.}
Figures~\ref{fig:boxplot-time-loadimages}, \ref{fig:boxplot-time-train}, \ref{fig:boxplot-memory-loadimages} and \ref{fig:boxplot-memory-train} show cost in execution time and memory, averaged across 10 users' data, as computed on the three real devices, and with two use case scenarios per $FL$ round.
First, we note that all costs reported are, on average, similar through the various $FL$ rounds.
This means that a device joining any of the $FL$ rounds is expected to have similar cost regardless if it is the beginning of the $FL$ process or later. 
Second, we find that execution in the newer devices, D1 and D2, is significantly faster than the older device D3, and this is true for both image loading, and especially model training.
Third, and expected, doubling the sample size as required in the second use case scenario (joint modeling), has the same execution cost per image loaded, but practically doubles the execution cost for model training.
Forth, we observe that the average memory cost for loading images is same across devices, regardless of scenario.
Fifth, and expected, when the sample size doubles ($S$=500), the model training consumes about double the memory than for $S$=250.
Finally, we measure the cost of data sharing between apps for materializing the second use case scenario.
We find that sending 250(500) samples between apps takes an extra time of 54.7(65.9)ms, and consumes 1.1(1.6)MBs of memory, on average, demonstrating the viability and low cost of this option for joint $FL$ modeling between apps on the same device.

\noindent
\textbf{FLaaS On-device Power \& CPU Costs.}
We measured power consumption and CPU utilization as estimated by the Battery Historian~\cite{batteryhistorian} tool (plots omitted due to space).
The mean CPU utilization per round for S=250(500) is 16.6(27.0)\%, 18.1(34.0)\% and 31.1(33.3)\%, for the 3 devices, respectively.
Interestingly, for S=250(500), D1 and D2 have similar average power consumption of 75.6(123.2) and 81.0(138.0)mAh, respectively, per $FL$ round.
Surprisingly, D3 has 10-12x higher power consumption than D1 or D2, of 924.8(1563.4)mAh.
This result is a consequence of the longer execution time and higher CPU utilization of D3 in comparison to D1 and D2, indicating that older devices may not be suitable for FL (or generally ML) model training.

\vspace{-4pt}
\section{Conclusions and Discussion}
In this paper, we presented \flaas, the first to our knowledge Federated Learning as a Service system enabling 3rd-party applications to build collaborative, decentralized, privacy-preserving ML models.
We discussed challenges arising under these settings, and highlighted approaches that can be used to solve them.
We also presented a \flaas proof of concept under mobile phone settings, showing the feasibility of our design and potential benefits of collaborative model building.
Our long-term goal is to finalize \flaas design and deployment for large scale evaluation.
However, we argue that many of the challenges raised in the \flaas design represent fundamental open research problems in the Federated Learning space.

\section*{Acknowledgements}
The research leading to these results has received funding from EU H2020 Programme, No 830927 (CONCORDIA), No 871793 (ACCORDION) and No 871370 (project PIMCITY).
The paper reflects only the authors' views and the Commission is not responsible for any use that may be made of the information it contains.

\newpage
\balance
\bibliographystyle{ACM-Reference-Format}
\begin{small}
\bibliography{references}


\begin{thebibliography}{00}


\ifx \showCODEN    \undefined \def \showCODEN     #1{\unskip}     \fi
\ifx \showDOI      \undefined \def \showDOI       #1{#1}\fi
\ifx \showISBNx    \undefined \def \showISBNx     #1{\unskip}     \fi
\ifx \showISBNxiii \undefined \def \showISBNxiii  #1{\unskip}     \fi
\ifx \showISSN     \undefined \def \showISSN      #1{\unskip}     \fi
\ifx \showLCCN     \undefined \def \showLCCN      #1{\unskip}     \fi
\ifx \shownote     \undefined \def \shownote      #1{#1}          \fi
\ifx \showarticletitle \undefined \def \showarticletitle #1{#1}   \fi
\ifx \showURL      \undefined \def \showURL       {\relax}        \fi
\providecommand\bibfield[2]{#2}
\providecommand\bibinfo[2]{#2}
\providecommand\natexlab[1]{#1}
\providecommand\showeprint[2][]{arXiv:#2}

\bibitem[\protect\citeauthoryear{Amazon}{Amazon}{2020}]%
        {aws-mlaas2020}
\bibfield{author}{\bibinfo{person}{Amazon}.} \bibinfo{year}{2020}\natexlab{}.
\newblock \bibinfo{title}{Machine Learning on AWS}.
\newblock
  \bibinfo{howpublished}{\url{https://aws.amazon.com/machine-learning/}}.
  (\bibinfo{year}{2020}).
\newblock


\bibitem[\protect\citeauthoryear{Android}{Android}{2020a}]%
        {broadcastreceiver}
\bibfield{author}{\bibinfo{person}{Android}.} \bibinfo{year}{2020}\natexlab{a}.
\newblock \bibinfo{title}{Android BroadcastReceiver}.
\newblock
  \bibinfo{howpublished}{\url{https://developer.android.com/reference/android/content/BroadcastReceiver}}.
    (\bibinfo{year}{2020}).
\newblock


\bibitem[\protect\citeauthoryear{Android}{Android}{2020b}]%
        {batteryhistorian}
\bibfield{author}{\bibinfo{person}{Android}.} \bibinfo{year}{2020}\natexlab{b}.
\newblock \bibinfo{title}{Android BroadcastReceiver}.
\newblock
  \bibinfo{howpublished}{\url{https://developer.android.com/topic/performance/power/setup-battery-historian}}.
    (\bibinfo{year}{2020}).
\newblock


\bibitem[\protect\citeauthoryear{Bellet, Guerraoui, Taziki, and Tommasi}{Bellet
  et~al\mbox{.}}{2018}]%
        {bellet2018p2p-ml}
\bibfield{author}{\bibinfo{person}{Aurelien Bellet}, \bibinfo{person}{Rachid
  Guerraoui}, \bibinfo{person}{Mahsa Taziki}, {and} \bibinfo{person}{Marc
  Tommasi}.} \bibinfo{year}{2018}\natexlab{}.
\newblock \showarticletitle{Personalized and Private Peer-to-Peer Machine
  Learning}. In \bibinfo{booktitle}{{\em 21st International Conference on
  Artificial Intelligence and Statistics (AISTATS)}}.
\newblock


\bibitem[\protect\citeauthoryear{Bonawitz, Eichner, Grieskamp, Huba, Ingerman,
  Ivanov, Kiddon, Konecny, Mazzocchi, McMahan, Overveldt, Petrou, Ramage, and
  Roselander}{Bonawitz et~al\mbox{.}}{2019}]%
        {bonawitz2019FL-sysml}
\bibfield{author}{\bibinfo{person}{Keith Bonawitz}, \bibinfo{person}{Hubert
  Eichner}, \bibinfo{person}{Wolfgang Grieskamp}, \bibinfo{person}{Dzmitry
  Huba}, \bibinfo{person}{Alex Ingerman}, \bibinfo{person}{Vladimir Ivanov},
  \bibinfo{person}{Chloe Kiddon}, \bibinfo{person}{Jakub Konecny},
  \bibinfo{person}{Stefano Mazzocchi}, \bibinfo{person}{H. McMahan},
  \bibinfo{person}{Timon Overveldt}, \bibinfo{person}{David Petrou},
  \bibinfo{person}{Daniel Ramage}, {and} \bibinfo{person}{Jason Roselander}.}
  \bibinfo{year}{2019}\natexlab{}.
\newblock \showarticletitle{Towards Federated Learning at Scale: System
  Design}. In \bibinfo{booktitle}{{\em 2nd SysML Conference}}.
\newblock


\bibitem[\protect\citeauthoryear{Briggs, Fan, and Andras}{Briggs
  et~al\mbox{.}}{2020}]%
        {briggs2020hierarchical-FL}
\bibfield{author}{\bibinfo{person}{Christopher Briggs}, \bibinfo{person}{Zhong
  Fan}, {and} \bibinfo{person}{Peter Andras}.} \bibinfo{year}{2020}\natexlab{}.
\newblock \bibinfo{title}{Federated learning with hierarchical clustering of
  local updates to improve training on non-IID data}.
\newblock \bibinfo{howpublished}{\url{https://arxiv.org/pdf/2004.11791.pdf}}.
  (\bibinfo{year}{2020}).
\newblock


\bibitem[\protect\citeauthoryear{Caldas, Meher Karthik~Duddu, Wu, Li, Konecny,
  McMahan, Smith, and Talwalkar}{Caldas et~al\mbox{.}}{2019}]%
        {leaf-benchmark2019}
\bibfield{author}{\bibinfo{person}{Sebastian Caldas}, \bibinfo{person}{Sai
  Meher Karthik~Duddu}, \bibinfo{person}{Peter Wu}, \bibinfo{person}{Tian Li},
  \bibinfo{person}{Jakub Konecny}, \bibinfo{person}{H.~Brendan McMahan},
  \bibinfo{person}{Virginia Smith}, {and} \bibinfo{person}{Ameet Talwalkar}.}
  \bibinfo{year}{2019}\natexlab{}.
\newblock \bibinfo{title}{LEAF: A Benchmark for Federated Settings}.
\newblock \bibinfo{howpublished}{\url{https://leaf.cmu.edu}}.
  (\bibinfo{year}{2019}).
\newblock


\bibitem[\protect\citeauthoryear{CoMind}{CoMind}{2019}]%
        {comind2019}
\bibfield{author}{\bibinfo{person}{CoMind}.} \bibinfo{year}{2019}\natexlab{}.
\newblock \bibinfo{title}{coMind: Collaborative Machine Learning}.
\newblock \bibinfo{howpublished}{\url{https://comind.org}}.
  (\bibinfo{year}{2019}).
\newblock


\bibitem[\protect\citeauthoryear{Datafleets}{Datafleets}{2020}]%
        {datafleets2020}
\bibfield{author}{\bibinfo{person}{Datafleets}.}
  \bibinfo{year}{2020}\natexlab{}.
\newblock \bibinfo{title}{Datafleets: The Federated Intelligence Platform}.
\newblock \bibinfo{howpublished}{\url{https://www.datafleets.com}}.
  (\bibinfo{year}{2020}).
\newblock


\bibitem[\protect\citeauthoryear{Deng, Dong, Socher, Li, Li, and Fei-Fei}{Deng
  et~al\mbox{.}}{2009}]%
        {imagenet_cvpr09}
\bibfield{author}{\bibinfo{person}{J. Deng}, \bibinfo{person}{W. Dong},
  \bibinfo{person}{R. Socher}, \bibinfo{person}{L.-J. Li}, \bibinfo{person}{K.
  Li}, {and} \bibinfo{person}{L. Fei-Fei}.} \bibinfo{year}{2009}\natexlab{}.
\newblock \showarticletitle{{ImageNet: A Large-Scale Hierarchical Image
  Database}}. In \bibinfo{booktitle}{{\em CVPR09}}.
\newblock


\bibitem[\protect\citeauthoryear{DML}{DML}{2018}]%
        {DML2018}
\bibfield{author}{\bibinfo{person}{DML}.} \bibinfo{year}{2018}\natexlab{}.
\newblock \bibinfo{title}{Decentralized Machine Learning}.
\newblock \bibinfo{howpublished}{\url{https://decentralizedml.com}}.
  (\bibinfo{year}{2018}).
\newblock


\bibitem[\protect\citeauthoryear{FATE}{FATE}{2020}]%
        {fate2020}
\bibfield{author}{\bibinfo{person}{FATE}.} \bibinfo{year}{2020}\natexlab{}.
\newblock \bibinfo{title}{Federated AI Technology Enabler}.
\newblock \bibinfo{howpublished}{\url{https://github.com/FederatedAI/FATE}}.
  (\bibinfo{year}{2020}).
\newblock


\bibitem[\protect\citeauthoryear{Geyer, Klein, and Nabi}{Geyer
  et~al\mbox{.}}{2017}]%
        {geyer2017FL-DP-central}
\bibfield{author}{\bibinfo{person}{Robin~C Geyer}, \bibinfo{person}{Tassilo
  Klein}, {and} \bibinfo{person}{Moin Nabi}.} \bibinfo{year}{2017}\natexlab{}.
\newblock \showarticletitle{Differentially private federated learning: A client
  level perspective}. In \bibinfo{booktitle}{{\em NIPS Workshop: Machine
  Learning on the Phone and other Consumer Devices}}.
\newblock


\bibitem[\protect\citeauthoryear{Google}{Google}{2020a}]%
        {googlecloud-mlaas2020}
\bibfield{author}{\bibinfo{person}{Google}.} \bibinfo{year}{2020}\natexlab{a}.
\newblock \bibinfo{title}{AI Platform}.
\newblock \bibinfo{howpublished}{\url{https://cloud.google.com/ai-platform}}.
  (\bibinfo{year}{2020}).
\newblock


\bibitem[\protect\citeauthoryear{Google}{Google}{2020b}]%
        {transfer-google}
\bibfield{author}{\bibinfo{person}{Google}.} \bibinfo{year}{2020}\natexlab{b}.
\newblock \bibinfo{title}{Google Transfer API library}.
\newblock
  \bibinfo{howpublished}{\url{https://github.com/tensorflow/examples/blob/master/lite/examples/model_personalization/}}.
    (\bibinfo{year}{2020}).
\newblock


\bibitem[\protect\citeauthoryear{Guha~Roy, Siddiqui, Polsterl, Navab, and
  Wachinger}{Guha~Roy et~al\mbox{.}}{2019}]%
        {roy2019brain-torrent}
\bibfield{author}{\bibinfo{person}{Abhijit Guha~Roy}, \bibinfo{person}{Shayan
  Siddiqui}, \bibinfo{person}{Sebastian Polsterl}, \bibinfo{person}{Nassir
  Navab}, {and} \bibinfo{person}{Christian Wachinger}.}
  \bibinfo{year}{2019}\natexlab{}.
\newblock \bibinfo{title}{BrainTorrent: A Peer-to-Peer Environment for
  Decentralized Federated Learning}.
\newblock \bibinfo{howpublished}{\url{https://arxiv.org/pdf/1905.06731.pdf}}.
  (\bibinfo{year}{2019}).
\newblock


\bibitem[\protect\citeauthoryear{Konecny, McMahan, Yu, Richtarik,
  Theertha~Suresh, and Bacon}{Konecny et~al\mbox{.}}{2016}]%
        {konecny2016federated-learning}
\bibfield{author}{\bibinfo{person}{Jakub Konecny}, \bibinfo{person}{H.~Brendan
  McMahan}, \bibinfo{person}{Felix Yu}, \bibinfo{person}{Peter Richtarik},
  \bibinfo{person}{Ananda Theertha~Suresh}, {and} \bibinfo{person}{Dave
  Bacon}.} \bibinfo{year}{2016}\natexlab{}.
\newblock \showarticletitle{Federated Learning: Strategies for Improving
  Communication Efficiency}. In \bibinfo{booktitle}{{\em 29th Conference on
  Neural Information Processing Systems(NIPS)}}.
\newblock


\bibitem[\protect\citeauthoryear{Krizhevsky, Hinton, et~al\mbox{.}}{Krizhevsky
  et~al\mbox{.}}{2009}]%
        {cifar10}
\bibfield{author}{\bibinfo{person}{Alex Krizhevsky}, \bibinfo{person}{Geoffrey
  Hinton}, {et~al\mbox{.}}} \bibinfo{year}{2009}\natexlab{}.
\newblock \showarticletitle{Learning multiple layers of features from tiny
  images}.
\newblock  (\bibinfo{year}{2009}).
\newblock


\bibitem[\protect\citeauthoryear{Liu, Zhang, Song, and Letaief}{Liu
  et~al\mbox{.}}{2020}]%
        {liu2019cloud-edge-FL}
\bibfield{author}{\bibinfo{person}{Lumin Liu}, \bibinfo{person}{Jun Zhang},
  \bibinfo{person}{S.H. Song}, {and} \bibinfo{person}{Khaled~B. Letaief}.}
  \bibinfo{year}{2020}\natexlab{}.
\newblock \showarticletitle{Client-Edge-Cloud Hierarchical Federated Learning}.
  In \bibinfo{booktitle}{{\em IEEE International Conference on
  Communications}}.
\newblock


\bibitem[\protect\citeauthoryear{McMahan, Moore, Ramage, and Ag{\"{u}}era~y
  Arcas}{McMahan et~al\mbox{.}}{2017}]%
        {mcmahan2017-FL}
\bibfield{author}{\bibinfo{person}{H.~Brendan McMahan}, \bibinfo{person}{Eider
  Moore}, \bibinfo{person}{Daniel Ramage}, {and} \bibinfo{person}{Blaise
  Ag{\"{u}}era~y Arcas}.} \bibinfo{year}{2017}\natexlab{}.
\newblock \showarticletitle{Communication-Efficient Learning of Deep Networks
  from Decentralized Data}. In \bibinfo{booktitle}{{\em 20th International
  Conference on Artificial Intelligence and Statistics (AISTATS)}}.
\newblock


\bibitem[\protect\citeauthoryear{Microsoft}{Microsoft}{2020}]%
        {azure-mlaas2020}
\bibfield{author}{\bibinfo{person}{Microsoft}.}
  \bibinfo{year}{2020}\natexlab{}.
\newblock \bibinfo{title}{Azure Machine Learning}.
\newblock
  \bibinfo{howpublished}{\url{https://azure.microsoft.com/en-us/services/machine-learning/}}.
    (\bibinfo{year}{2020}).
\newblock


\bibitem[\protect\citeauthoryear{OpenMined}{OpenMined}{2020}]%
        {openmined2020}
\bibfield{author}{\bibinfo{person}{OpenMined}.}
  \bibinfo{year}{2020}\natexlab{}.
\newblock \bibinfo{title}{OpenMined}.
\newblock \bibinfo{howpublished}{\url{https://www.openmined.org}}.
  (\bibinfo{year}{2020}).
\newblock


\bibitem[\protect\citeauthoryear{Salehi Heydar~Abad, Ozfatura, Gunduz, and
  Ercetin}{Salehi Heydar~Abad et~al\mbox{.}}{2020}]%
        {mehdi2019hierarchical-FL-cellular}
\bibfield{author}{\bibinfo{person}{Mehdi Salehi Heydar~Abad},
  \bibinfo{person}{Emre Ozfatura}, \bibinfo{person}{Deniz Gunduz}, {and}
  \bibinfo{person}{Ozgur Ercetin}.} \bibinfo{year}{2020}\natexlab{}.
\newblock \bibinfo{title}{Hierarchical Federated Learning Across Heterogeneous
  Cellular Networks}.
\newblock   (\bibinfo{year}{2020}).
\newblock


\bibitem[\protect\citeauthoryear{Sandler, Howard, Zhu, Zhmoginov, and
  Chen}{Sandler et~al\mbox{.}}{2018}]%
        {s2018mobilenetv2}
\bibfield{author}{\bibinfo{person}{Mark Sandler}, \bibinfo{person}{Andrew
  Howard}, \bibinfo{person}{Menglong Zhu}, \bibinfo{person}{Andrey Zhmoginov},
  {and} \bibinfo{person}{Liang-Chieh Chen}.} \bibinfo{year}{2018}\natexlab{}.
\newblock \showarticletitle{MobileNetV2: Inverted Residuals and Linear
  Bottlenecks}. In \bibinfo{booktitle}{{\em {IEEE/CVF Conference on Computer
  Vision and Pattern Recognition}}}.
\newblock


\bibitem[\protect\citeauthoryear{Seif, Tandon, and Li}{Seif
  et~al\mbox{.}}{2020}]%
        {seif2020wireless-LDP-FL}
\bibfield{author}{\bibinfo{person}{Mohamed Seif}, \bibinfo{person}{Ravi
  Tandon}, {and} \bibinfo{person}{Ming Li}.} \bibinfo{year}{2020}\natexlab{}.
\newblock \showarticletitle{Wireless federated learning with local differential
  privacy}. In \bibinfo{booktitle}{{\em {IEEE International Symposium on
  Information Theory (ISIT)}}}.
\newblock


\bibitem[\protect\citeauthoryear{Tan, Sun, Kong, Zhang, Yang, and Liu}{Tan
  et~al\mbox{.}}{2018}]%
        {tan2018DL-transfer-learning-survey}
\bibfield{author}{\bibinfo{person}{Chuanqi Tan}, \bibinfo{person}{Fuchun Sun},
  \bibinfo{person}{Tao Kong}, \bibinfo{person}{Wenchang Zhang},
  \bibinfo{person}{Chao Yang}, {and} \bibinfo{person}{Chunfang Liu}.}
  \bibinfo{year}{2018}\natexlab{}.
\newblock \showarticletitle{A Survey on Deep Transfer Learning}. In
  \bibinfo{booktitle}{{\em Artificial Neural Networks and Machine Learning
  (ICANN)}}, \bibfield{editor}{\bibinfo{person}{V{\v{e}}ra K{\r{u}}rkov{\'a}},
  \bibinfo{person}{Yannis Manolopoulos}, \bibinfo{person}{Barbara Hammer},
  \bibinfo{person}{Lazaros Iliadis}, {and} \bibinfo{person}{Ilias
  Maglogiannis}} (Eds.). \bibinfo{publisher}{Springer International
  Publishing}, \bibinfo{address}{Cham}, \bibinfo{pages}{270--279}.
\newblock


\bibitem[\protect\citeauthoryear{TensorFlow}{TensorFlow}{2020}]%
        {tensorflow-federated2020}
\bibfield{author}{\bibinfo{person}{TensorFlow}.}
  \bibinfo{year}{2020}\natexlab{}.
\newblock \bibinfo{title}{TensorFlow Federated: Machine Learning on
  Decentralized Data}.
\newblock \bibinfo{howpublished}{\url{https://www.tensorflow.org/federated}}.
  (\bibinfo{year}{2020}).
\newblock


\bibitem[\protect\citeauthoryear{Truex, Baracaldo, Anwar, Steinke, Ludwig,
  Zhang, and Zhou}{Truex et~al\mbox{.}}{2019}]%
        {truex2019hybrid-FL}
\bibfield{author}{\bibinfo{person}{Stacey Truex}, \bibinfo{person}{Nathalie
  Baracaldo}, \bibinfo{person}{Ali Anwar}, \bibinfo{person}{Thomas Steinke},
  \bibinfo{person}{Heiko Ludwig}, \bibinfo{person}{Rui Zhang}, {and}
  \bibinfo{person}{Yi Zhou}.} \bibinfo{year}{2019}\natexlab{}.
\newblock \showarticletitle{A hybrid approach to privacy-preserving federated
  learning}. In \bibinfo{booktitle}{{\em Proceedings of the 12th ACM Workshop
  on Artificial Intelligence and Security}}.
\newblock


\bibitem[\protect\citeauthoryear{Truex, Liu, Chow, Gursoy, and Wei}{Truex
  et~al\mbox{.}}{2020}]%
        {truex2020LDP-FED}
\bibfield{author}{\bibinfo{person}{Stacey Truex}, \bibinfo{person}{Ling Liu},
  \bibinfo{person}{Ka-Ho Chow}, \bibinfo{person}{Mehmet~Emre Gursoy}, {and}
  \bibinfo{person}{Wenqi Wei}.} \bibinfo{year}{2020}\natexlab{}.
\newblock \showarticletitle{LDP-Fed: Federated Learning with Local Differential
  Privacy}. In \bibinfo{booktitle}{{\em 3rd International Workshop on Edge
  Systems, Analytics and Networking (EdgeSys)}}.
\newblock


\bibitem[\protect\citeauthoryear{Wei, Li, Ding, Ma, Yang, Farokhi, Jin, Quek,
  and Poor}{Wei et~al\mbox{.}}{2020}]%
        {wei2020FL-DP-local}
\bibfield{author}{\bibinfo{person}{Kang Wei}, \bibinfo{person}{Jun Li},
  \bibinfo{person}{Ming Ding}, \bibinfo{person}{Chuan Ma},
  \bibinfo{person}{Howard~H Yang}, \bibinfo{person}{Farhad Farokhi},
  \bibinfo{person}{Shi Jin}, \bibinfo{person}{Tony~QS Quek}, {and}
  \bibinfo{person}{H~Vincent Poor}.} \bibinfo{year}{2020}\natexlab{}.
\newblock \showarticletitle{Federated learning with differential privacy:
  Algorithms and performance analysis}.
\newblock \bibinfo{journal}{{\em IEEE Transactions on Information Forensics and
  Security\/}} (\bibinfo{year}{2020}).
\newblock


\bibitem[\protect\citeauthoryear{Zhang, Wang, Zhao, and Chen}{Zhang
  et~al\mbox{.}}{2019}]%
        {zhang2019efficient-FL-DP}
\bibfield{author}{\bibinfo{person}{Jiale Zhang}, \bibinfo{person}{Junyu Wang},
  \bibinfo{person}{Yanchao Zhao}, {and} \bibinfo{person}{Bing Chen}.}
  \bibinfo{year}{2019}\natexlab{}.
\newblock \showarticletitle{An Efficient Federated Learning Scheme with
  Differential Privacy in Mobile Edge Computing}. In \bibinfo{booktitle}{{\em
  International Conference on Machine Learning and Intelligent
  Communications}}. Springer, \bibinfo{pages}{538--550}.
\newblock


\bibitem[\protect\citeauthoryear{Zhao, Kaafar, and Kourtellis}{Zhao
  et~al\mbox{.}}{2020a}]%
        {zhao2020privacy-tradeoff}
\bibfield{author}{\bibinfo{person}{Benjamin Zi~Hao Zhao},
  \bibinfo{person}{Mohamed~Ali Kaafar}, {and} \bibinfo{person}{Nicolas
  Kourtellis}.} \bibinfo{year}{2020}\natexlab{a}.
\newblock \showarticletitle{{Not one but many Tradeoffs: Privacy Vs. Utility in
  Differentially Private Machine Learning}}. In \bibinfo{booktitle}{{\em {ACM
  Cloud Computing Security Workshop (CCSW)}}}.
\newblock


\bibitem[\protect\citeauthoryear{Zhao, Zhao, Yang, Wang, Wang, Lyu, Niyato, and
  Lam}{Zhao et~al\mbox{.}}{2020b}]%
        {zhao2020LDP-FL-vehicles}
\bibfield{author}{\bibinfo{person}{Yang Zhao}, \bibinfo{person}{Jun Zhao},
  \bibinfo{person}{Mengmeng Yang}, \bibinfo{person}{Teng Wang},
  \bibinfo{person}{Ning Wang}, \bibinfo{person}{Lingjuan Lyu},
  \bibinfo{person}{Dusit Niyato}, {and} \bibinfo{person}{Kwok-Yan Lam}.}
  \bibinfo{year}{2020}\natexlab{b}.
\newblock \showarticletitle{Local Differential Privacy based Federated Learning
  for Internet of Things}.
\newblock \bibinfo{journal}{{\em EEE Internet of Things Journal\/}}
  (\bibinfo{year}{2020}).
\newblock


\bibitem[\protect\citeauthoryear{Zhu, Kairouz, McMahan, Sun, and Li}{Zhu
  et~al\mbox{.}}{2020}]%
        {zhu2020heavyhitters}
\bibfield{author}{\bibinfo{person}{Wennan Zhu}, \bibinfo{person}{Peter
  Kairouz}, \bibinfo{person}{Brendan McMahan}, \bibinfo{person}{Haicheng Sun},
  {and} \bibinfo{person}{Wei Li}.} \bibinfo{year}{2020}\natexlab{}.
\newblock \showarticletitle{Federated Heavy Hitters Discovery with Differential
  Privacy}. In \bibinfo{booktitle}{{\em {23 International Conference on
  Artificial Intelligence and Statistics (AISTATS)}}}.
\newblock


\end{thebibliography}
\end{small}

\end{document}